\documentclass[twocolumn]{ceurart}
\makeatletter
\DeclareOldFontCommand{\rm}{\normalfont\rmfamily}{\mathrm}
\DeclareOldFontCommand{\sf}{\normalfont\sffamily}{\mathsf}
\DeclareOldFontCommand{\tt}{\normalfont\ttfamily}{\mathtt}
\DeclareOldFontCommand{\bf}{\normalfont\bfseries}{\mathbf}
\DeclareOldFontCommand{\it}{\normalfont\itshape}{\mathit}
\DeclareOldFontCommand{\sl}{\normalfont\slshape}{\@nomath\sl}
\DeclareOldFontCommand{\sc}{\normalfont\scshape}{\@nomath\sc}
\makeatother

\usepackage{graphicx}
\usepackage{dirtytalk}
\usepackage{xcolor}
\usepackage{hyperref}
\usepackage{amsmath}
\usepackage{amssymb}
\usepackage{bbold}
\usepackage{microtype}
\usepackage{booktabs} 
\usepackage[utf8]{inputenc}
\usepackage{caption}
\usepackage{subcaption}

\usepackage{pbox}

\begin{document}

\copyrightyear{2020}
\copyrightclause{Copyright for this paper by its authors.
  Use permitted under Creative Commons License Attribution 4.0
  International (CC BY 4.0).}

\conference{Proceedings of the CIKM 2020 Workshops, October 19-20, 2020, Galway, Ireland}

\title{Matching the Clinical Reality: Accurate OCT-Based Diagnosis From Few Labels}

\author[1,2]{Valentyn Melnychuk}[orcid=0000-0002-2401-6803]
\ead{v.melnychuk@campus.lmu.de}
\address[1]{Fraunhofer Institute for Integrated Circuits IIS, Erlangen, Germany}
\address[2]{Ludwig Maximilian University of Munich, Germany}

\author[2]{Evgeniy Faerman}
\ead{faerman@dbs.ifi.lmu.de}

\author[2]{Ilja Manakov}
\ead{ilja.manakov@gmx.de}

\author[2]{Thomas Seidl}[orcid=0000-0002-4861-1412]
\ead{seidl@dbs.ifi.lmu.de}

\begin{abstract}
    Unlabeled data is often abundant in the clinic, making machine learning methods based on semi-supervised learning a good match for this setting. Despite this, they are currently receiving relatively little attention in medical image analysis literature.
    Instead, most practitioners and researchers focus on supervised or transfer learning approaches. The recently proposed MixMatch and FixMatch algorithms have demonstrated promising results in extracting useful representations while requiring very few labels.
    Motivated by these recent successes, we apply MixMatch and FixMatch in an ophthalmological diagnostic setting and investigate how they fare against standard transfer learning. We find that both algorithms outperform the transfer learning baseline on all fractions of labelled data. Furthermore, our experiments show that exponential moving average (EMA) of model parameters, which is a component of both algorithms, is not needed for our classification problem, as disabling it leaves the outcome unchanged. Our code is available online: \href{https://github.com/Valentyn1997/oct-diagn-semi-supervised}{https://github.com/Valentyn1997/oct-diagn-semi-supervised}.
\end{abstract}


\maketitle

\section{Introduction}
In recent years deep learning techniques have taken the field of AI by storm.
Virtually all state-of-the-art systems in computer vision (CV) rely on some form of deep learning.
This paradigm shift has sparked the imagination of many practitioners and researchers in the medical image analysis domain.
Computer-aided diagnosis appeared to be next-in-line to benefit from the advancements made in CV, as the amount of data in clinical diagnostics is increasing rapidly.
The research community has proposed a plethora of new algorithms and systems for the automated diagnosis of a wide range of diseases.
However, clinical adoption has been slow. One crucial reason is that supervised learning, which forms the basis for the vast majority of deep learning approaches, is ill-suited to the medical domain. 

This mismatch is two-fold. For one, the labelled data needed for supervised learning is prohibitively costly to generate for medical applications.
With a shortage of medical practitioners, diverting medical experts' time and energy to labelling efforts becomes exceedingly expensive.
More fine-grained problem formulations (e.g. single-label vs. multi-label, volume level vs. slice level annotation, etc.) result in exponentially more labelling expenses.
Additionally, most clinics lack the tools to label vast amounts of data. Secondly, and perhaps more fundamentally, there is an epistemic problem in generating accurate labels.
For any given diagnostic problem, the inter-expert agreement is well below 100\%.
This discrepancy stems from the fact that medicine is complex and does not always fit neatly into a classification formulation.
Additionally, each expert comes with his or her own set of experiences and knowledge.

Instead of solely relying on supervised learning, semi-supervised learning (SSL) should discover the bulk \sloppy of the knowledge required for solving a diagnostic task on its own, with labels only serving as additional guidance. The idea of SSL is to train a machine learning algorithm on vast amounts of unlabeled data and a small set of labelled samples. 
SSL is a much better match for the clinical setting, as unlabeled data is often abundant since it is acquired as part of the clinical routine.

In this work, we apply two recently proposed SSL methods, MixMatch \cite{berthelot2019mixmatch} and FixMatch \cite{Sohn2020FixMatchSS}, to a diagnostic problem in ophthalmology.
We test which performs better in classifying optical coherence tomography (OCT) b-scans into four classes (one healthy and three pathological) at different fractions of labelled data.
We compare the two SSL methods to a baseline transfer learning approach, similar to \cite{kermany2018identifying}.
After going over related work in the next section, we explain the basis for our experiments in Section \ref{sec:methods}, covering MixMatch, FixMatch and the transfer learning baseline. In Section \ref{sec:experiments} we describe the dataset and present the results of our investigation. We conclude with Section \ref{sec:conclusion} by summarizing our findings and discussing how they apply to the clinical setting.
    
\section{Related work}

\paragraph{\textbf{Semi-supervised learning}.} State-of-the-art methods for image classification concentrate on finding the right combination of SSL paradigms. One of the early approaches -- Mean-Teacher \cite{tarvainen2017mean} -- uses exponential moving average (EMA) of model parameters. Virtual Adversarial Training (VAT) \cite{miyato2018virtual}, tries to find a minimal perturbation and fit a robust model against it. MixMatch \cite{berthelot2019mixmatch} and RealMix \cite{nair2019realmix} encompass mixing and overlaying labelled and unlabelled images to obtain consistent predictions. Unsupervised Data Augmentation (UDA) \cite{Xie2019UnsupervisedDA} uses strongly augmented images to force consistency among unlabeled images. ReMixMatch \cite{berthelot2019remixmatch} uses so-called \say{augmentation anchoring}, i.e. strong and weak augmentations, to enforce consistency. Inspired by UDA and ReMixMatch, the authors of FixMatch \cite{Sohn2020FixMatchSS} significantly simplify SSL by relying only on augmentations and pseudo-labelling with a confidence threshold. We provide a broader overview of applied SSL methods in Section \ref{subsec:SSL}. 

Surprisingly, there exists only a little amount of literature on SSL applied to ophthalmological data. \cite{liu2018seg} and \cite{sedai2019seg} utilize SSL for OCT segmentation. 
%\cite{liu2018seg} leverages a discriminator network trained on unlabeled instances to refine segmentation maps. In contrast, \cite{sedai2019seg} trains a teacher network on the labelled data and uses dropout to obtain an uncertainty map of the segmentation. They then proceed to label additional samples using the teacher network and train a student network using the uncertainty map to discount labels from the teacher network.
In the domain of automated diagnosis, \cite{liu2019semi} employ an autoencoder with an additional classification module on the latent code in the detection of retinopathy from colour fundus images.
%For unlabeled instances the network receives a supervision signal form the reconstruction criterion, while labelled datapoints generate an additional signal through the classification module.
\cite{xie2019semi} tackle the same problem by extending the GAN framework \cite{goodfellow2014gan} to one \say{fake} and six \say{real} classes, i.e. the labeled classes.
Recent works \cite{he2020retinal} and \cite{das2020data} apply the same principle to the classification of OCT b-scans.
%The discriminator is the classifier in this work and unlabeled data is leveraged by requiring the discriminator to classify them as one of the \say{real} classes.
Most recently, \cite{wang2020semi} applied SSL methods to glaucoma detection by imputing missing visual field (VF) measurements through nearest-neighbour identification in the latent space of a pre-trained classification CNN.
Afterwards, \cite{wang2020semi} train a multi-task network jointly on glaucoma classification and VF measurement prediction.
To the best of our knowledge, we are the first to apply consistency regularization based SSL techniques (see Section \ref{sec:methods}) to the problem of automated diagnosis in ophthalmology.

\paragraph{\textbf{Transfer learning}.} Among numerous approaches existing in the deep transfer learning \cite{tan2018survey}, we choose the fine-tuning or network-based transfer learning to be the most promising. 
%Namely, this is done by fully or partially reusing the network weights, pre-trained on huge labelled image dataset, in other domain. 
\cite{oquab2014learning, donahue2014decaf} proposed to use ImageNet \cite{imagenetcvpr09} pre-trained CNN as the initialization for different visual recognition tasks with the limited amount of labels. Yosinski et al. \cite{yosinski2014transferable} discovered how unfreezing different parts of the network while fine-tuning affects the target performance. 

% \paragraph{Transfer learning.} Among numerous approaches existing in the deep transfer learning \cite{tan2018survey}, we choose the fine-tuning or network-based transfer learning to be the most promising. Namely, this is done by fully or partially reusing the network weights, pre-trained on huge labelled image dataset, in other domain. \cite{oquab2014learning, donahue2014decaf} proposed to use ImageNet \cite{imagenetcvpr09} pre-trained deep convolutional neural network as the initialisation for different visual recognition tasks with the limited amount of labels. Yosinski et al. \cite{yosinski2014transferable} discovered how unfreezing different parts of the network while fine-tuning affects the target performance. 

% \paragraph{\textbf{Transfer Learning with Semi-Supervised Learning}.} Authors of \cite{zhou2018semi} experimented with using pre-trained models as the initialization for SSL algorithms and concluded that this gives barely any improvement.

% \paragraph{\textbf{Transfer learning with Self-Supervised Learning}.} Recently proposed self-supervised methods, such as SimCLR \cite{chen2020simple} and SimCLRv2 \cite{Chen2020BigSM} over-perform SSL methods described above on the ImageNet 10\% labeled data benchmark. They were pre-trained in the unsupervised fashion using contrastive loss and then fine-tuned with the few labelled samples. We find those methods extremely computation-intensive, as they require a huge batch size. 
    
\section{Approach} \label{sec:methods}
Transfer learning and semi-supervised learning are two main approaches for predictive modelling when dealing with data with few labels. Transfer learning approaches reuse knowledge from previously learned tasks. On the other hand, the SSL approaches allow learning with small labelled datasets by utilizing unlabeled data from the same distribution in the learning process. In the following, we first discuss our transfer learning baseline and afterwards describe the SSL approaches we have chosen for this study. 

\subsection{Transfer Learning} \label{subsec:transfer-learning}
When applying transfer learning techniques, the user has to choose how to adapt the model from the auxiliary to the primary task.
In our experiments, we use a network, which was pre-trained on ImageNet \cite{imagenetcvpr09}.
For adapting the model to OCT classification we try two common approaches.
In the \emph{feature extraction} approach, we freeze all parameters except for the final fully connected layer, analogous to \cite{kermany2018identifying}.
Alternatively, we use the pre-trained network as initialization and allow all parameters to change. We refer to this as \emph{fine-tuning} hereafter.

% Transfer learning utilizes only labelled data in the training process. In our work, we experiment with
% two settings, cf. e.g., \cite{sabatelli2018deep}:
% %two different approaches for pre-trained CNN fine-tuning (similarly to ): using pre-trained networks as feature extractor and full network fine-tuning. 

% \paragraph{\textbf{Feature extraction}.} 
% %One freezes all pre-trained CNN layers except the last, meaning
% The last classification layer of the pre-trained CNN is replaced by a new one. The pre-trained weights of the network stay the same, only the last layer is trained. Therefore the neural network plays the role of the feature extractor and the problem is simplified to the multi-class logistic regression.
% This approach tries to overcome the problem of overfitting since, only a small fraction of parameters is optimized, making this approach better suited to a setting with few labels.  

% \paragraph{\textbf{Fine-tuning}.} The pre-trained CNN is used for the initialization of the weights of the new network. The whole new network is learned for the new problem.

\subsection{Semi-supervised Learning} \label{subsec:SSL}
In our study we compare two of recent state-of-the-art algorithms for SSL \emph{MixMatch} \cite{berthelot2019mixmatch} and \emph{FixMatch} \cite{Sohn2020FixMatchSS}. Both algorithms combine several pre-existing techniques from SSL. In this chapter, we review the main ideas and compare their utilization in both algorithms. We refer the reader to Appendix \ref{app:algorithms} for the detailed algorithm descriptions.
\paragraph{\textbf{Data Augmentation.}}
Data Augmentation is a regularization technique which is often used in supervised learning. The goal is that the model's prediction is not affected by the certain transformation of data instances. Therefore additional training data is added to the dataset by applying various perturbations to the data while keeping original labels. Most of the data augmentations are domain-specific and require domain knowledge.

\emph{MixMatch} uses random flip-and-shift augmentations (horizontal flips and random crops) for both labelled and unlabeled data.

\emph{FixMatch} distinguishes between \emph{weak} and \emph{strong} \sloppy data augmentations. Flip-and-shift augmentations are considered as weak augmentations, whereas affine trasformations and color-jittering are examples of strong augmentations (originally -- 14 different transformations from RandAugment \cite{Cubuk2019RandAugmentPA}). 

\paragraph{Pseudo-Labelling.}
Pseudo-Labeling or self-training loss \cite{lee2013pseudo} is the process of using the trained model to obtain labels for unlabeled instances. The predicted labels are used to guide the further learning process, e.g. by using generated labels as new targets.

\emph{MixMatch} applies different augmentations for an unlabeled instance and computes the class distribution for each augmentation. Therefore, instead of \emph{hard} one hot label MixMatch defines a probability distribution as the target. To sharpen the distribution and to reduce its entropy, the temperature of distribution is adjusted \cite{hinton2015distilling}.

\emph{FixMatch} uses a \say{classic} version of pseudo-labelling with hard labels and fixed confidence. The class probability distribution is taken from model outputs after a weak augmentation. If the probability of the most probable class exceeds a predefined threshold the label is assigned to a strongly augmented version of the same instance and used in the loss calculation.

\paragraph{Consistency Regularization.} Consistency regularization \cite{sajjadi2016regularization} imposes the constraint that the model should \sloppy make similar predictions for the same instance under different data augmentations.
Both \emph{MixMatch} and \emph{FixMatch} apply data augmentation on labelled and unlabeled data and enforce similar prediction for the same instance under different augmentations. For the unlabeled instances, the pseudo-label is used as a target. \emph{FixMatch} uses soft augmentations to compute pseudo-labels for hard augmentations of the same training sample.

\paragraph{EMA decay.}
Another popular consistency requirement in SSL is an exponential moving average (EMA) of model parameters over time to smooth the behaviour when the model changes its decisions rapidly. The EMA is the basis of Mean Teacher algorithm \cite{tarvainen2017mean}, which maintains two models. The \emph{teacher} model stores an exponential moving average of \emph{student's} parameters and is used to make the predictions to compute the pseudo-labels. Therefore pseudo-labels computed by the teacher can be considered as a weighted combination of decisions of previous models.
The \emph{student} model makes the predictions for the training data and is updated based on the training loss.
Both \emph{MixMatch} and \emph{FixMatch} employ EMA decay while inference.
Note, that keeping a second model in memory and updating its parameters results in higher memory requirements and computation costs.

\paragraph{MixUp.}
MixUp \cite{zhang2017mixup} is another regularization technique to avoid overfitting. MixUp linearly combines training instance pairs and their prediction targets. Therefore it tries to impose linear behaviour between training samples.
\emph{MixMatch} does not differentiate between \emph{pseudo} targets predicted for the unlabeled instances and \emph{ground truth} labels and mixes all possible target pairs. Therefore a resulting instance used in training may be a combination of two pseudo targets, two ground truth labels or of pseudo-target with ground truth label.

%Two recent SOTA approaches were taken to the consideration in this study: MixMatch \cite{berthelot2019mixmatch} and FixMatch \cite{Sohn2020FixMatchSS}. Both methods exploit a specific version of \textbf{consistency regularization} -- an idea that classifier should predict the same label for both labelled and unlabelled image under random augmentations.

\section{Experiments} \label{sec:experiments}
Our work follows the principles of the fair SSL evaluation framework, defined by \cite{oliver2018realistic}. The authors highlight the importance of using the same classifying model structure for comparison. The evaluation is also meaningful for the real use-case if SSL methods are compared with well-fine-tuned transfer learning and fully supervised models.
%Also one needs to always compare SSL methods with both well-fine-tuned full-supervised and transfer learning models. Additionally, we use a realistically small validation subset and check the model performances on the varying amount of labelled and unlabelled data.

\begin{figure*}
    \centering
    \begin{subfigure}{.75\textwidth}
        \centering
        \includegraphics[width=\linewidth]{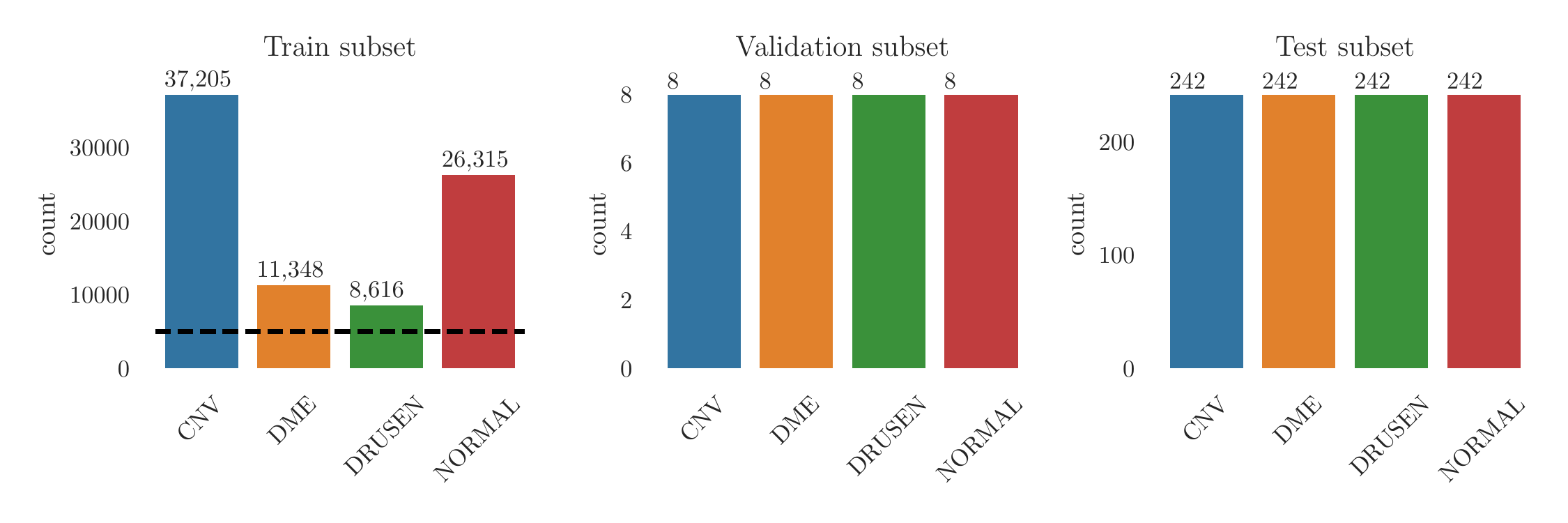}
    \end{subfigure}
    % \hspace{-0.5cm}
    \begin{subfigure}{.24\textwidth}
        \centering
        \includegraphics[width=\linewidth]{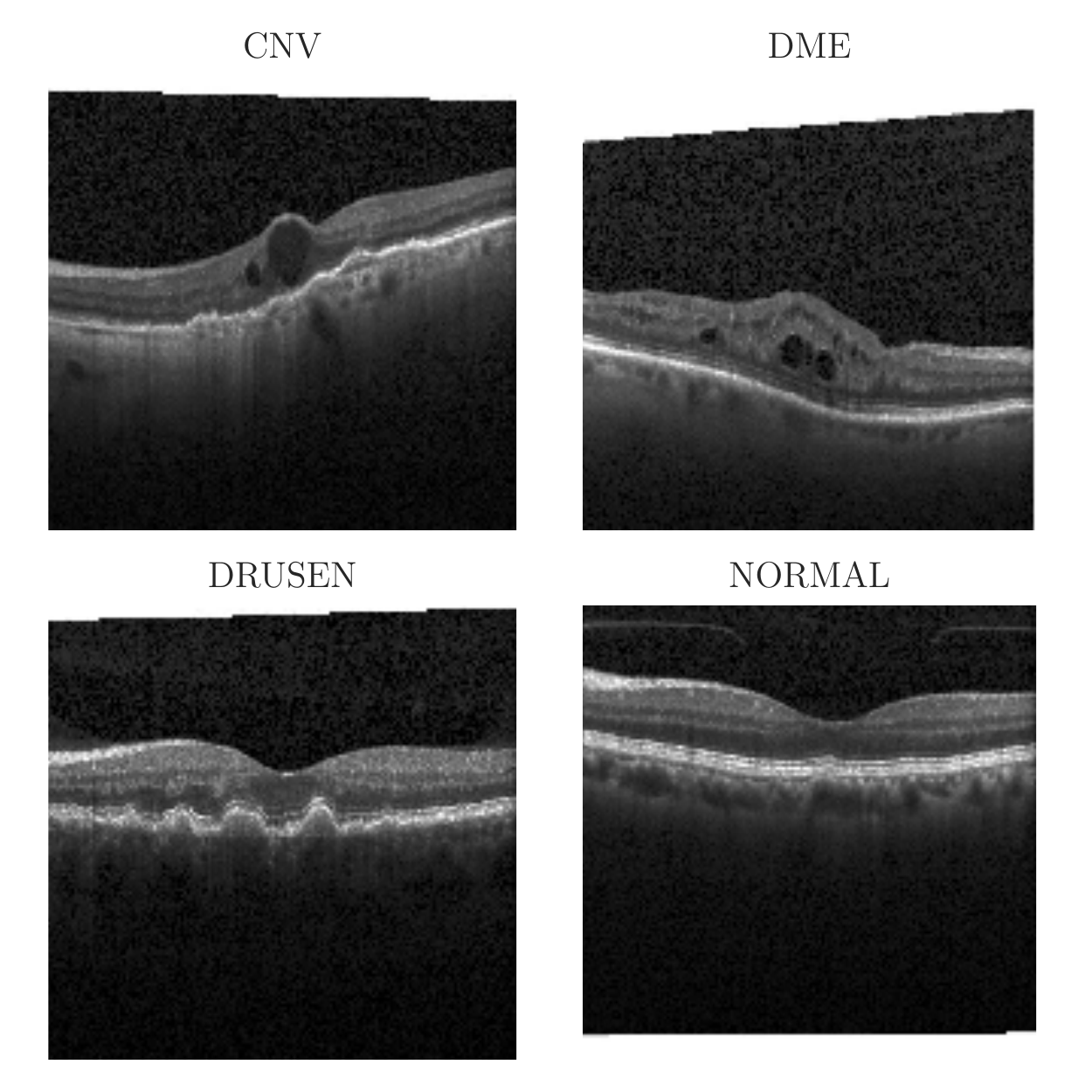}
    \end{subfigure}
    \caption{Histogramms of image labels and random image for each class. Horizontal dashed line on train subset subplot denotes labelled-unlabelled split with $n_l = 20,000$ (upper part corresponds to unlabelled subset). The images on the right depict a sample from each class.}
    \label{fig:dataset_dist}
\end{figure*}

For the evaluation we use the \textbf{UCSD dataset} published by Kermany et al. \cite{kermany2018identifying}. It contains 84,495 optical coherence tomography (OCT) b-scans pertaining to four categories; \say{normal}, \say{drusenoid} (DRUSEN), \say{choroidal neovascularization} (CNV) and \say{diabetic macular edema} (DME).
The images vary in size, where the median image has a size of 496$\times$512 pixels. The height of the images ranges between 496 and 512 and the width between 384 and 1536. 
%For all models we apply the same preprocessing with basic transformations, $\alpha(\cdot)$ augmentations, resizing to 512$\times$512 pixels, the random square crop with scale range 0.5-1.0 and resizing to 256$\times$256 pixels.
%Images are monochrome with heights varying in range 496-512 pixels and widths -- 384-1536 pixels (median image has a dimensionality of 496$\times$512). We turned monochrome images to RGB by duplicating the channel three times. Basic transformations and augmentation $\alpha(\cdot)$ for train subset are subsequent resize to 512$\times$512 pixels, horizontal flip, random square crop with scale range 0.5-1.0 and resize to 256$\times$256 pixels. Test and validation images are resized to 256$\times$256 pixels.
The dataset is also obtainable through Kaggle\footnote{\href{https://www.kaggle.com/paultimothymooney/kermany2018}{https://www.kaggle.com/paultimothymooney/kermany2018}.}. For better comparability, we use the same train/validation/test split as in the Kaggle challenge. There are several images for each patient in the dataset and splits are done patient-wise, there are no images of the same patients in different splits. Test and validation are balanced, there are 8 and 242 images per class respectively (see Fig. \ref{fig:dataset_dist}).
In our experiments, we vary the number of labelled data, which we sample randomly from the training subset. We sample the same number of labelled training instances from each class. For SSL approaches the rest of the train set is used as unlabeled data. 

%Dataset of OCT images OCT2017  has 4 classes -- normal and 3 types of anomalies and. It was split in train, validation and test subsets (we consider a Kaggle version), validation and test being balanced (8 and 242 images per class respectively) (see Figure \ref{fig:dataset_dist}). Dataset split ensures, that both test and train+validation subsets contain no studies of the same patient. To split train subset into labelled and unlabelled parts we simply take first $n_l$ images and drop all the remaining labels, so that the labelled subset is balanced. Due to balanced test subset accuracy is chosen as the main performance metric.

We compare the performance of transfer learning and SSL models using the same \textbf{Wide ResNet-50-2} \cite{DBLP:journals/corr/ZagoruykoK16} backbone. Since the images are monochrome we duplicate the channel three times for RGB channels.

%\sloppy Finally, we report accuracy for varying number of labels $n_l \in \{40, 100, 200, 800, 2,000, 4,000, 20,000\}$ (see Figure \ref{fig:best_models}). Dashed line depicts the test accuracy of fully-supervised Wide ResNet-50-2, trained from scratch with all the 83,484 labelled train images and using the same augmentations. It was not fine-tuned and should be considered an approximate upper baseline for transfer and semi-supervised learning.

%We also experiment with the Mean Teacher technique \cite{tarvainen2017mean} (see Section \ref{sec:methods}). We tried several values for the decay parameter of the EMA $\beta_{\text{EMA}}$ ranging from 0.0 (no EMA) to 0.999 (slow update). EMA model is updated after each batch. 

For each model, we perform hyperparameter search, described in Appendix \ref{app:transfer-learning} and \ref{app:SSL}. For all experiments, we report the model performance on test data in the epoch with the lowest validation loss.
%the test accuracy of the epochs with the lowest validation loss.

\subsection{Comparison of transfer learning and SSL approaches}
First, in Table \ref{table:reported-results} we compare the performance of our backbone model trained with all labelled instances to the results reported previously in the literature for the same UCSD dataset. As we can see, the backbone model achieves almost perfect performance when trained with enough labels.

\begin{figure}
    \centering
    \begin{subfigure}{.45\textwidth}
        \centering
        \includegraphics[width=\linewidth]{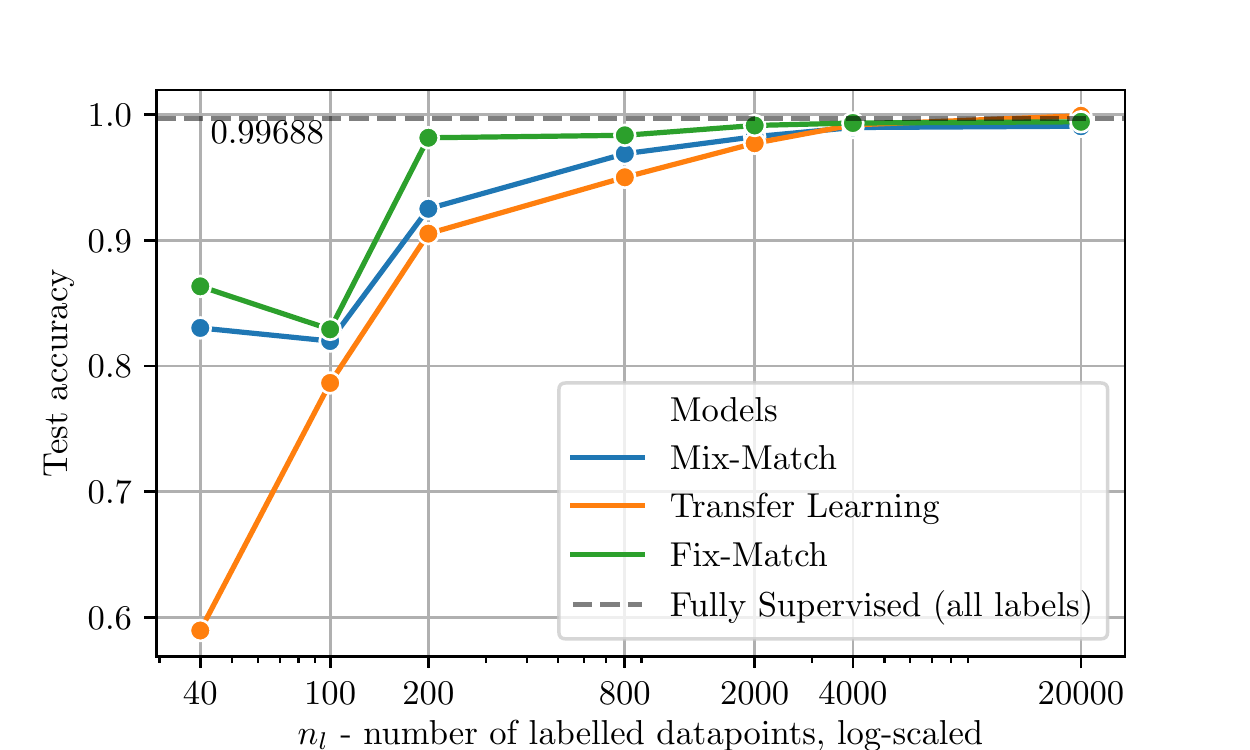}
        \caption{Best models, maximum performance among 4 runs (Semi-supervised) / 8 runs (Transfer learning) per each $n_l$}
        \label{fig:best_models}
    \end{subfigure}
    \hspace{0.07cm}
    \begin{subfigure}{.45\textwidth}
        \centering
        \includegraphics[width=\linewidth]{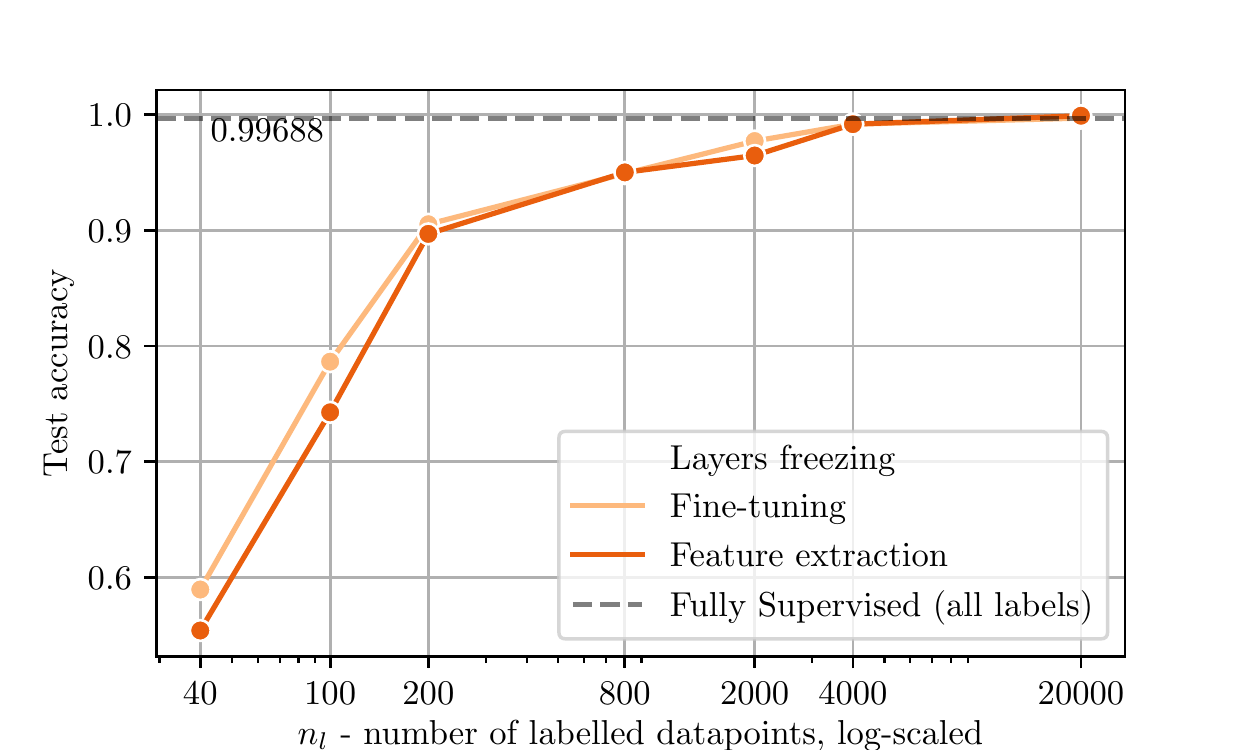}
        \caption{Study of layers freezing for Transfer learning, maximum performance among 4 runs per each $n_l$.}
        \label{fig:freezing}
    \end{subfigure}
    \caption{Test accuracies for SSL and Transfer learning models for varying number of labels $n_l$. Fully-supervised baseline with all labels uses EMA decay  (${\beta_{\text{EMA}}} = 0.999$).}
    \label{fig:models_comparison}
\end{figure}

Next, in Fig. \ref{fig:freezing} we compare two transfer learning approaches. Note, that the hyperparameter search was done for each number of labels for each approach.
We discover that, contrary to our expectations, the fine-tuning variant outperforms feature extraction approach in all label settings. We believe that a thorough selection of hyperparameters with representative validation set reduces the risk of overfitting. Furthermore, since the original models are trained on the dataset with RGB channels, we believe that the model can better adapt to the monochrome setting when all model weights are allowed to be changed.

In the Fig. \ref{fig:best_models} we present the results of both SSL algorithms and compare them with the best performing transfer learning setting. We find that the SSL approaches outperform transfer learning on all fractions of labelled data. The gap between SSL and transfer learning widens significantly for smaller fractions of labelled data.  With only 10 labelled representatives per class, the FixMatch achieves an accuracy of over 86\%, while transfer learning reaches only 59\%. We also see, that with about 2000-4000 labels all methods achieve almost perfect performance. The \emph{Fix-Match} algorithm also outperforms  \emph{Mix-Match} in almost all settings and with only 50 labelled points per class achieves the accuracy of 98.14\%. We also observe a small SSL performance drop for 25 labelled images per class -- mainly because methods require even more epochs to fit (we employ a heuristical formula for defining the maximum number of epochs based on the number of labels, see Appendix \ref{app:SSL}, \ref{eq:number-epochs}).    
%We see that both semi-supervised methods perform better in the sparse labels setting (Figure \ref{fig:best_models} (Left) and Table \ref{tab:ema_comparison}). FixMatch seems to perform the best for all $n_l$, except $n_l=20,000$, where Transfer Learning (99.90\%) outperforms FixMatch (99.41\%) by a small margin.  
%The small drop in the overall trend of FixMatch is also noticeable for $n_l=100$, probably due to a few $n_B$ for this setting or a badly chosen labelled subset.
%Regarding training time (see Table \ref{tab:training_time}), time requirements are inversely proportional to the models' performances:
%\begin{itemize}
%    \item Transfer learning ignores unlabelled data, overfits fastly.
%    \item MixMatch and FixMatch usually require much time to converge (MixMatch was also overfitting). They use more images on each data-step. 
%\end{itemize}

\begin{table}
    \centering
    \setlength{\tabcolsep}{2pt}
    \renewcommand{\arraystretch}{1}
    \begin{tabular}{l|l|l|l}
         \toprule
         Method & $n_l$ & Accuracy & Notes \\ \midrule
         Kermany et al. \cite{kermany2018identifying} & All & 96.6\% & Original paper \\ \hline
         Alqudah \cite{alqudah2020aoct}
         & All & 97.1\% & \pbox{15cm}{\vspace{4pt} Extended UCSD \\ with 5 classes \vspace{4pt}}  \\ \hline
         Wu et al. \cite{wu2020attennet} & All & 97.5\% \\ \hline
         Chetoui et al. \cite{chetoui2020deep} & All & 98.46\% \\ \hline
         Tsuji et al.\cite{tsuji2020classification} & All & 99.6\% \\ \hline
         \pbox{10cm}{\vspace{4pt}\textbf{WideResNet-50-2} \\ \textbf{(our backbone)}} & All & \textbf{99.69\%} & \pbox{15cm}{\vspace{4pt} With EMA decay \\ (${\beta_{\text{EMA}}} = 0.999$) \vspace{4pt}}   \\ \midrule
%         & 100 & 86.29 $\pm$ 1.83\%\\ 
%         Das et al. \cite{das2020data} & 300 & 95.31 $\pm$ 1.18 \%\\
%          & 500 & 97.43 $\pm$ 0.66\%\\ \hline
          He et al. \cite{he2020retinal} & 835 & 87.25$\pm$1.44\% * & *Average precision \\
         \bottomrule
    \end{tabular}
    % \vspace{0.15cm}
    \caption{Reported test accuracies for UCSD dataset. Methods have different backbones and thus are not fully comparable with the proposed SLL methods. Nevertheless, our best fully-supervised model outperforms previously reported methods.}
    \label{table:reported-results}
\end{table}

Finally, since practitioners have often to deal with the resource constraints and actual running times are rarely reported in the literature, we report them in Table \ref{tab:training_time}. Note, that all methods are implemented in the same framework and the experiments are done on the same machine with two Tesla V100 Nvidia GPUs. To use the same batch size as recommended in the original publications, we have used both GPUs to train \emph{Fix-Match}. Other models are trained on a single GPU.

\begin{table*}
    \centering
    \setlength{\tabcolsep}{3pt}
    \renewcommand{\arraystretch}{1}
    \begin{tabular}{l||l|l|l|l|l|l|l}
    \toprule
    \multicolumn{1}{r||}{$n_l$} &       \multicolumn{1}{c|}{40}  &       \multicolumn{1}{c|}{100}   &    \multicolumn{1}{c|}{200}   &    \multicolumn{1}{c|}{800}   &     \multicolumn{1}{c|}{2000}  &    \multicolumn{1}{c|}{4000}  &    \multicolumn{1}{c}{20000} \\
    \midrule
    Transfer Learning &        10m  &         9m  &     12m  &     15m  &      24m  &     39m  &  1h 39m  \\
    Mix-Match         &  1d 16h 5m  &     9h 12m  &  6h 13m  &  2h 30m  &   2h 37m  &  2h 24m  &  2h 26m  \\
    FixMatch         &  5d 9h 36m  &  1d 19h 4m  &  1d 40m  &  9h 58m  &  10h 40m  &  9h 50m  &  7h 51m  \\
    \bottomrule
    \end{tabular}
    % \vspace{0.15cm}
    \caption{Training time comparison between the best models of each approach for varying number of labels $n_l$. We do not include the time, spent on hyperparameter search and report only the training time of single models.}
    \label{tab:training_time}
\end{table*}

%\subsection{Layers freezing}
%By comparison on Figure \ref{fig:best_models} (Right), there is no obvious advantage in full or FC layer backbone unfreezing. EMA decay ($\beta_{\text{EMA}}$) or learning rate parameters seem to play a greater role while fine-tuning in transfer learning.  We assume that could be related to a huge dimensionality of the last FC layer in Wide ResNet-50-2.

\subsection{EMA decay}
The \emph{EMA} is inherent part of \emph{Fix-Match} algorithm and is also optionally recommended for \emph{Mix-Match}. We observe learning curve to be more stable for validation subset for all the models when models are trained using it. However, we assume that with the right chosen validation subset, the variability could be advantageous and one can find a better fit. Usage of \emph{EMA} model causes additional computation and memory costs and as can be seen in Table \ref{tab:ema_comparison} most of the time models without it perform better.

%We notice no clear advantage of using EMA of model parameters (see Table \ref{tab:ema_comparison}), even though it was reported as an advancement for FixMatch or possible extension for MixMatch. Originally, $\beta_{\text{EMA}} = 0.999$ is dedicated to smooth the prediction of classifier by calculating the EMA of previous training parameters \cite{DBLP:journals/corr/TarvainenV17}. Indeed, we observe learning curves to be more stable for both train and validation subsets for all the models. However, we assume that with the right chosen validation subset, the variability of parameters actually could be advantageous and one can find a better fit.

\begin{table*}
    % \centering
    \setlength{\tabcolsep}{3pt}
    \renewcommand{\arraystretch}{1}
    \begin{tabular}{l||c c|c c|c c|c c|c c|c c|c c}
    \toprule
    \multicolumn{1}{r||}{$n_l$}  & \multicolumn{2}{c|}{40} & \multicolumn{2}{c|}{100} & \multicolumn{2}{c|}{200} & \multicolumn{2}{c|}{800} & \multicolumn{2}{c|}{2000} & \multicolumn{2}{c|}{4000} & \multicolumn{2}{c}{20000} \\
    \multicolumn{1}{r||}{${\beta_{\text{EMA}}}$} &    0.0 &  0.999 &    0.0 &  0.999 &    0.0 &  0.999 &    0.0 &  0.999 & 0.0 &  0.999 &    0.0 &  0.999 &    0.0 &  0.999 \\
    \midrule
    Transfer Learning &  58.96 &  --- &  78.65 &  --- &  90.52 &  --- &  95.00 &  --- & 97.71 &  --- &  98.23 &  --- &  99.38 &  --- \\
    Mix-Match         &  83.02 &  55.00 &  75.73 &  81.98 &  92.50 &  88.85 &  94.69 &  96.88 &  98.02 &  98.23 &  98.02 &  98.96 &  99.06 &  98.75 \\
    FixMatch         &  \textbf{86.33} &  72.66 &  \textbf{82.91} &  75.88 &  97.07 &  \textbf{98.14} &  \textbf{98.34} &  98.05 &  97.85 &  \textbf{99.12} &  98.34 &  \textbf{99.32} &  98.63 &  \textbf{99.41} \\
    \bottomrule
    \end{tabular}
    
    % \begin{tabular}{l||c c|c c|c c}
    % \toprule
    % \multicolumn{1}{r||}{${n_l}$}  & \multicolumn{2}{c|}{2000} & \multicolumn{2}{c|}{4000} & \multicolumn{2}{c}{20000} \\
    % \multicolumn{1}{r||}{$\mathbf{\beta_{\text{EMA}}}$} &    0.0 &  0.999 &    0.0 &  0.999 &    0.0 &  0.999  \\
    % \midrule
    % Transfer Learning &  97.71 &  --- &  98.23 &  --- &  99.38 &  --- \\
    % MixMatch         &  96.98 &  98.23 &  96.67 &  98.23 &  99.06 &  97.92 \\
    % FixMatch         &  97.85 &  \textbf{99.12} &  98.34 &  \textbf{99.32} &  98.63 &  \textbf{99.41} \\
    % \bottomrule
    % \end{tabular}
    
    % \vspace{0.15cm}
    \caption{Best test accuracy based on several runs (8 runs with varying learning rate, weight decay and back-bone unfreezing for Transfer Learning, 2 runs with a varying total number of batches for MixMatch and FixMatch) for varying number of labels $n_l$. We do not observe any profit of using / not using EMA decay (${\beta_{\text{EMA}}}$) for all methods, unlike previously reported results.}
    \label{tab:ema_comparison}
    
\end{table*}

\section{Conclusion} \label{sec:conclusion}
In this work, we have demonstrated the efficacy of MixMatch and FixMatch, when applied to an ophthalmological diagnostic problem on OCT data. The two algorithms were able to attain high accuracy, achieving well over 80\% on as little as 40 labelled samples (i.e. ten per class). Both algorithms outperformed transfer learning in the few labelled data settings.
This study emphasizes the use of SSL methods in the clinical adoption of AI.
Although both MixMatch and FixMatch are more computationally expensive than transfer learning, the amount of labelling effort saved by using them is immense.
With labelling being one of the biggest factors hindering clinical use of AI methodology, we argue that smarter use of the abundance of unlabeled data already present at the clinic will be a major strategy for overcoming this hurdle.

As part of future work, we propose to also compare SSL approach with the few-shot deep learning. 

\bibliography{bibliography}

%%
%% The acknowledgements section is defined using the "acks" environment
\begin{acknowledgments}
  This work has been funded by the German Federal Ministry of Education and Research (BMBF) under Grant No. 01IS18036A and by the Bavarian Ministry for Economic Affairs, Infrastructure, Transport and Technology through the Center for Analytics-Data-Applications (ADA-Center) within the framework of “BAYERN DIGITAL II”. The authors of this work take full responsibilities for its content.
\end{acknowledgments}

\appendix
\newpage
\section*{Appendix}

\section{MixMatch \& FixMatch -- algorithm details} \label{app:algorithms}

% The same backbone plays both teacher's and student's roles: as the teacher network generates pseudo-labels, the student back-propagates consistency loss based on real and pseudo-labels.
The foundation of both MixMatch and FixMatch is consistency regularization -- the idea that augmentations of the same data point should yield the same label. In this way, the model regularizes itself based on its predictions.

Let $\mathcal{X} = \{(x_b, y_b), b \in (1, ..., B) \}$ be the batch of labeled examples. $\alpha(\cdot)$ denotes the set of \emph{weak} augmentations and $\mathcal{A}(\cdot)$ -- \emph{strong} augmentations. $\hat{y} = f_{\text{M}}(x; \theta))$ is the prediction of backbone classifier, parametrized by $\theta$. $H(\cdot, \cdot)$ denotes categorical cross-entropy and $\lambda_u$ is unsupervised loss weight.

\textbf{MixMatch} employs only weak augmentations $\alpha(\cdot)$ and MixUp \cite{zhang2017mixup}. Let $\mathcal{U} = \{u_b, b \in (1, ..., B) \}$ be the unlabeled data batch. The model outputs of $K$ random weak augmentations $\alpha(\cdot)$ of the same unlabelled sample are treated as soft pseudo-labels $q_b$. These soft pseudo-labels are averaged and sharpened with the temperature $T$ for each image in $\mathcal{U}$ to yield a pseudo-label for that image. Then, images from both randomly augmented $\mathcal{X}$ and $\mathcal{U}$ are concatenated and shuffled, resulting in set $\mathcal{W}$. Afterwards, samples in $\mathcal{X}$ and $\mathcal{U}$ are weakly augmented and linearly interpolated with samples from $\mathcal{W}$. This results in $\hat{\mathcal{X}}$ and $\hat{\mathcal{U}}$ -- "mixed-up" versions of augmented labelled and $K$ unlabelled batches. Coefficients of MixUp are sampled from $\text{Beta}(\alpha, \alpha)$ distribution. The final loss is the sum of categorical cross-entropy for images from $\hat{\mathcal{X}}$ (supervised part) and Brier score for $\hat{\mathcal{U}}$ images (unsupervised part): 
\begin{multline}
    \mathcal{L} = \frac{1}{B}\sum_{(x, y) \in \hat{\mathcal{X}}} \operatorname{H}(y, f_{\text{M}}(x; \theta)) + \\ + \frac{\lambda_u}{KB} \sum_{(u, q) \in \hat{\mathcal{U}}} {||q - f_{\text{M}}(u; \theta)||}^2_2,
\end{multline}
MixMatch linearly ramps up $\lambda_u$ from 0 to its maximum after each batch to reduce the influence of unsupervised part during early stages of training.

\textbf{FixMatch} is a more simplified method. Unlabeled data batch $\mathcal{U} = \{u_b,  b \in (1, ..., \mu B) \}$ is now $\mu$-times bigger. Given the model's prediction $q_b$ for a weakly augmented unlabelled sample $u_b$, method yields hard pseudo-labels $\hat{q_b} = \operatorname{arg max}(q_b)$ and $\hat{\mathcal{U}} = \{(u_b, q_b, \hat{q_b})\}$. Afterwards, the model predicts labels for both a batch of weakly augmented labelled images and a batch of strongly augmented unlabelled images. Only the confident predictions for unlabelled samples are used in the final unsupervised part of the loss. They are filtered with the threshold $\tau$. The loss of FixMatch is then the sum of two categorical cross-entropies for labelled and unlabelled images:
\begin{multline}
     \mathcal{L} = \frac{1}{B}\sum_{(x, y) \in {\mathcal{X}}} \operatorname{H}(y, f_{\text{M}}(\alpha(x); \theta)) + \\ + \frac{\lambda_u}{\mu B} \sum_{(u, q, \hat{q}) \in \hat{\mathcal{U}}} \mathbb{1}_{\max(q) > \tau} \operatorname{H}(\hat{q},  f_{\text{M}}(\mathcal{A}(u); \theta))
\end{multline}
       
As we use $\tau$ for filtering confident pseudo-labels, we do not need the linear ramp-up for $\lambda_u$.

\section{Experiments}
\subsection{Transfer learning} \label{app:transfer-learning}
We took a version of Wide ResNet-50-2 pre-trained on ImageNet from PyTorch.\footnote{\href{https://pytorch.org/hub/pytorch_vision_wide_resnet/}{https://pytorch.org/hub/pytorch\_vision\_wide\_resnet/}.} Transfer learning was fine-tuned for every individual $n_l$, as it did not require much computational budget: 
\begin{itemize}
    \item learning rate $\in \{1*10^{-3}, 5*10^{-4}\}$
    \item optimizer weight decay $\in \{0.0, 0.0001 \}$
    \item layers freezing $\in \{$Fine-tuning, \\ Feature extraction$\}$ (see Section \ref{subsec:transfer-learning})
    % \item $\beta_{\text{EMA}} \in \{ 0.0, 0.999\}$
\end{itemize} 

Further hyperparameters are kept fixed, namely  we use Adam optimizer \cite{kingma2014adam}, $B = 32$, number of epochs = 50. Additionally, early stopping with the patience of 25 epochs was applied to avoid overfitting.

\subsection{MixMatch \& FixMatch} \label{app:SSL}

Hyperparameter fine-tuning for both SSL methods was two-fold: firstly, we fine-tuned more general parameters on 200 labelled samples ($n_l = 200$) with respect to the validation loss (see Table \ref{tab:grid_search}). Secondly, for each specific $n_l$, we tuned subset-size-dependent parameters. 

The labeled batch size was $B = 16$ for both algorithms. Additionally, we fix $\mu = 4, \tau = 0.7$ for FixMatch. We omit using cosine learning rate decay.  

\begin{table}
    \centering
    \setlength{\tabcolsep}{2.5pt}
    \renewcommand{\arraystretch}{1}
    \begin{tabular}{l|c|c}
        \toprule
         Hyperparameter &  MixMatch & FixMatch\\ \midrule
         Learning rate & $\{\mathbf{0.01}, 0.001\}$ & $\{\mathbf{0.03}\}$ \\
         Optimizer & $\{ \textbf{Adam} \}$ & $\{$Adam, $\textbf{SGD}\}$ \\
         Number of epochs & $\{500, \mathbf{1000}\}$ & $\{1000, \mathbf{2000}\}$ \\
         $T$ & $\{0.25, \mathbf{0.5}, 0.75, 0.9\}$ & --- \\
         $\alpha$ & $\{ 0.25, 0.5, 0.75, \mathbf{0.9} \}$ & --- \\
         $\lambda_u$ & $\{ 12.5, \mathbf{25}, 50, 100, 150 \}$ & $\{ \mathbf{5.0}, 25.0 \}$ \\ \midrule
         Grid-search size & 320 & 8 \\
         \bottomrule
    \end{tabular}
    \vspace{0.2cm}
    \caption{Primary hyperparameter search grid for SSL methods. Best value is marked with bold font. SGD -- stocastic gradient descent with momentum ($\beta = 0.9$) \cite{nesterov1983method}. An epoch is defined by maximum number of batches in labelled subset.}
    \label{tab:grid_search}
\end{table}
% \vspace{-1.5cm}

Regarding secondary fine-tuning, after the increase of $n_l$, each epoch becomes proportionally longer. Thus, we propose the following inverse formula to define the number of epochs:
\begin{equation} \label{eq:number-epochs}
    \text{Number of epochs} = \operatorname{round} \bigg( \frac{n_B}{n_l \operatorname{div} B} \bigg),
\end{equation}
where $n_B$ denotes total number of labelled batches, used while training. 

While secondary fine-tuning, we vary:
\begin{itemize}
    \item $\beta_{\text{EMA}} \in \{ 0.0, 0.999\}$ (EMA decay)
    \item $n_B \in \{12000, 15000\}$ for MixMatch / \\ $n_B \in \{24000, 30000\}$ for FixMatch
\end{itemize}

\end{document}